\documentclass[preprint,12pt,authoryear]{elsarticle}

\usepackage{amssymb}
\usepackage[authoryear]{natbib}
\usepackage{subcaption}

\usepackage{tikz}
\usepackage{float}
\usepackage[edges]{forest}
\usepackage[T1]{fontenc}
\usepackage{lmodern}  
\usepackage[table]{xcolor}
\usepackage{makecell}
\usepackage{multirow}
\usepackage{booktabs}
\usepackage{amsmath}
\usepackage{url}
\usepackage{gensymb}
\usepackage{graphicx}
\usepackage{verbatim}
\usepackage{tabularx} % Add this in the preamble if not already included
\usepackage{algorithm}
\usepackage{algorithmicx}
\usepackage{algpseudocode}
\usepackage{minted}
\usepackage{listings}
\usepackage[utf8]{inputenc}
\RequirePackage{stfloats}
\usepackage{xurl}
\usepackage[colorlinks=true, linkcolor=blue, citecolor=blue, urlcolor=blue, hidelinks]{hyperref} 
\usepackage{adjustbox}
\usepackage{siunitx}

\def\tsc#1{\csdef{#1}{\textsc{\lowercase{#1}}\xspace}}
\tsc{WGM}
\tsc{QE}
\begin{document}

\begin{frontmatter}

% ------------------------------------------------
% TITLE
% ------------------------------------------------

\title{Visual-SLAM for the detection of hidden tomatoes in greenhouses by Hierarchical Localization and GLOMAP for robotized harvesting}

% ------------------------------------------------
% AUTHORS
% ------------------------------------------------

\author[UAL-Inf]{Fernando Cañadas-Aránega}\corref{corr}\ead{fernando.ca@ual.es}
\cortext[corr]{Corresponding author}

\author[UAL-Inf]{José C. Moreno}\ead{jcmoreno@ual.es}

\author[UAL-Eng]{José L. Blanco-Claraco}\ead{jlblanco@ual.es}

\author[UAL-Inf]{F. Rodríguez}\ead{frrodrig@ual.es}

% ------------------------------------------------
% AFFILIATIONS
% ------------------------------------------------

\affiliation[UAL-Inf]{
    organization={Department of Informatics, CIESOL, ceiA3, Universidad de Almería},
    addressline={Ctra. Sacramento s/n},
    city={Almería},
    postcode={04120},
    country={Spain}
}

\affiliation[UAL-Eng]{
    organization={Department of Engineering, CIESOL, ceiA3, Universidad de Almería},
    addressline={Ctra. Sacramento s/n},
    city={Almería},
    postcode={04120},
    country={Spain}
}

% ------------------------------------------------
% ABSTRACT
% ------------------------------------------------

\begin{abstract}
Advanced crop monitoring inside greenhouses is becoming one of the primary objectives of research centers. High-performance sensors, such as LiDAR or stereo cameras, have traditionally been employed for this purpose, though these often have a high cost. This work proposes a Visual-SLAM system using a monocular camera, which is significantly more cost-effective and specifically tailored for agricultural applications, such as mapping tomato crops in a greenhouse. Tests were carried out on a real tomato bunch, located in the Agroconnect experimental greenhouse. A ROS 2 Humble node was developed to run on the robot in order to capture images of these crops, which were then stored for offline processing. To generate a 3D mapped model for the crop in the greenhouse, the GLOMAP mapper, based on Structure-From-Motion, was integrated with the Hierarchical Localization toolbox. This initial mapping is a foundation for future, more advanced algorithms to analyze growth patterns, and optimize agricultural management. The system leverages a hierarchical localization paradigm based on a coarse-to-fine strategy: it first performs global retrieval to generate location hypotheses, then combines local features within the identified candidate regions. The results show a correct identification of the tomato cluster, correctly characterising the tomato that is occluded and inaccessible by classical vision technologies. The reconstructed 3D model was further validated against manual ground-truth measurements of fruit size, centroid position, and orientation, confirming the geometric accuracy of the proposed low-cost monocular pipeline.
\end{abstract}

% ------------------------------------------------
% KEYWORDS
% ------------------------------------------------


\end{frontmatter}

% Main text
\section{Introduction}

Greenhouses exhibit a certain degree of structural organization but differ significantly from highly controlled industrial environments, such as automotive assembly lines. This distinction presents specific challenges that require the integration of highly automated machinery to ensure the efficient development of greenhouse agriculture. A key aspect in the deployment of robotic systems in such environments is the selection of appropriate sensors capable of detecting obstacles and localizing the robot in dynamic settings with limited communication \citep{feng2015design,canadas2026integrated}. These systems must not only navigate autonomously but also perform complex tasks such as automated crop harvesting. 

Maximizing the efficient use of available space within the greenhouse becomes a critical strategy for achieving optimal productivity, which in turn demands the adaptation of task-specific algorithms to the environmental conditions. Research on robotic tasks in greenhouses has been ongoing since 1987 \citep{bachche2015deliberation}. It is estimated that up to 80\% of operational time during the crop season is devoted to monitoring and fruit harvesting, the latter being one of the most studied tasks by the authors \citep{taqi2017cherry}. The most successful developments have incorporated robotic arms for automated harvesting, typically equipped with Red, Green and Blue – Depth (RGB-D) cameras for shape and depth detection, and/or LiDAR sensors for 3D scanning and environmental mapping \citep{canadas2024autonomous}, or by using RGB-D cameras to carry out ground segmentation, detection or classification \citep{canadas2026greenseg}.

Accurate environmental mapping is essential to determine the 3D position of fruits, which are then processed by object detection and classification algorithms. Using traditional techniques with 2D cameras has been a failure because the information obtained from this type of sensor is insufficient to identify the fruit \citep{ge2021yolox}. However, there are strategies that aim to improve these algorithms using RGB-D cameras to obtain three-dimensional models that can locate the position of the object in space. In \cite{rong2022fruit}, the YOLOv5 algorithm \cite{ge2021yolox} is used to detect tomato clusters from 2D images and estimate their position. In \cite{zheng2024fruit} the YOLOv5\_se variant is employed to identify tomatoes using RGB and depth images. Similarly, \cite{aranega2025nbv} applies SLAM techniques to estimate a 3D model of tomatoes using 3D LiDAR technology. However, accurately determining the total number of fruits remains a challenge, as many are occluded and can not be detected by these technologies. This limitation has driven the development of algorithms capable of estimating occlusions and improving the efficiency of automated harvesting \citep{aguilar2024preliminary}.

In this context, the three-dimensional image restructuring algorithms from motion, also known as Visual-SLAM, are a possible tool to be investigated for robotized crop monitoring in agriculture. In particular, the Hierarchical Localization (HLoc) \citep{sarlin2019from}, an algorithm that provides feature matching between consecutive images with a SuperPoint + SuperGlue structure. On the other hand, the Global Structure-from-Motion Revisited (GLOMAP) algorithm \citep{pan2024global} produces larger-scale reconstructions—approximately one to two times bigger—while incorporating features specifically designed to enhance depth perception in the resulting 3D models.

This paper presents an application of work of Pablo Vela \citep{vela2025hloc}, which introduces a repository for utilizing the GLOMAP Structure-From-Motion mapper in conjunction with the Hierarchical Localization toolbox to achieve a faster mapper with depth-learned features and matches. Initially, a rosbag is recorded using ROS2 Humble on a real tomato cluster located in a real greenhouse. For the recording, an Intel RealSense D435i camera is used, from which only the information recorded by the RGB lens will be utilized. This information is converted to a frame rate of 10 Hz, which will be used as input to the algorithm. Subsequently, HLoc-GLOMAP is used for the active reconstruction of the cluster. The technique yields promising results, demonstrating how a set of 2D images can be used to reconstruct a real 3D model, a key aspect for locating a percentage of the objects and planning a more efficient future harvesting strategy. The main contribution of this work is the identification of the tomatoes that are hidden in the plant for the subsequent planning of harvesting, in order to speed up the harvesting problems typical of this type of vegetables, from a sequence of 2D images, which in general can be acquired by a low cost camera.

This paper is organized as follows: Section 2 describes the materials and methods that have been used. Section 3 presents the results of applying this technique to a real data set. Finally, Section 4 outlines the main conclusions of this research.

\section{Materials and methods} \label{sec: 2}

This subsection describes the materials and methods employed in this paper. 

\subsection{Materials}

\subsubsection{Agroconnect facilities}

\begin{figure}[H]
  \centering
  \begin{subfigure}{0.9\linewidth} \centering
    \includegraphics[width=10cm]{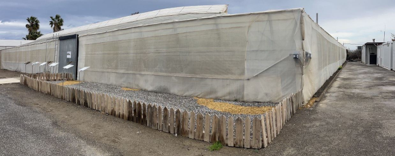} \centering
    \caption{Exterior}
    \label{fig:sub11}
  \end{subfigure}
  %\hspace{1cm}
  \begin{subfigure}{0.9\linewidth} \centering
    \includegraphics[width=10cm]{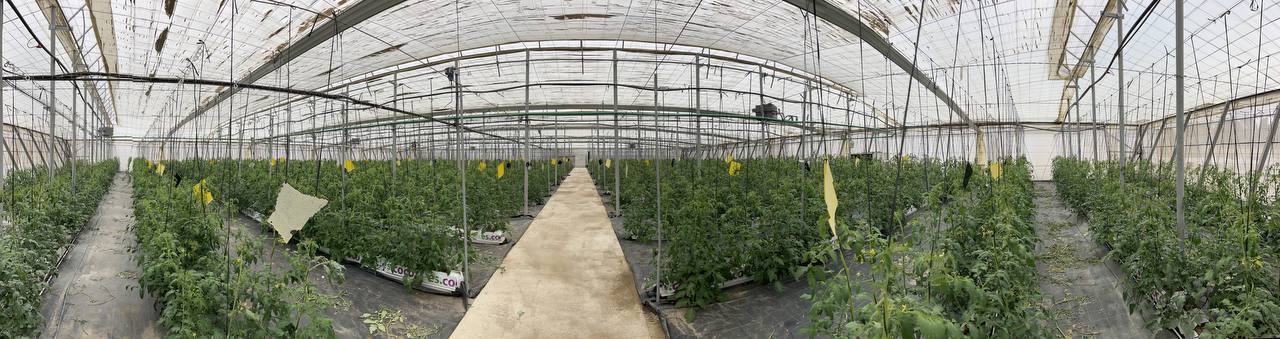} \centering
    \caption{Interior}
    \label{fig:sub12}
  \end{subfigure}
  \caption{AgroConnect experimental greenhouse} \label{fig:Fig11}
\end{figure}

The experimental trials were conducted at the Agroconnect facilities described in \citep{moreno2022modelado}, located in the municipal district of La Cañada de San Urbano (Almería, Spain), at 36°50' N, 2°24' W, with an elevation of 3 meters above sea level and a terrain slope of 1\% toward the north and a surface area of 1850 m$^2$ (see Figure \ref{fig:Fig11}). These facilities have been co-funded by the Ministry of Science, Innovation and Universities, in collaboration with the European Regional Development Fund (FEDER), through the 2019 grant program for the acquisition of advanced scientific and technological infrastructure. This greenhouse has been the subject of numerous research projects carried out by the research group, ranging from the development of a datasheet incorporating LiDAR sensors, cameras and IMUs \citep{canadas2024multimodal} to the development of advanced control techniques applied to mobile robots \citep{canadas2026adaptive}. In addition, a benchmark is available that focuses on testing control algorithms in this same greenhouse \citep{canadas2026ros2}.

\begin{figure}[H]
  \centering
  \begin{subfigure}{0.45\linewidth} \centering
    \includegraphics[width=7.15cm]{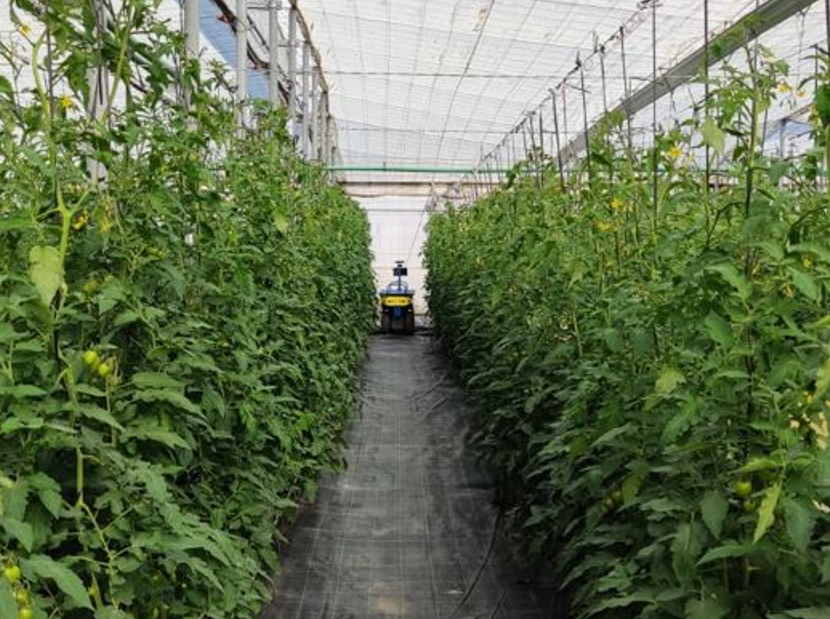} \centering
    \caption{Robot between plants}
    \label{fig:sub1}
  \end{subfigure}
  %\hspace{1cm}
  \begin{subfigure}{0.45\linewidth} \centering
    \includegraphics[width=4cm]{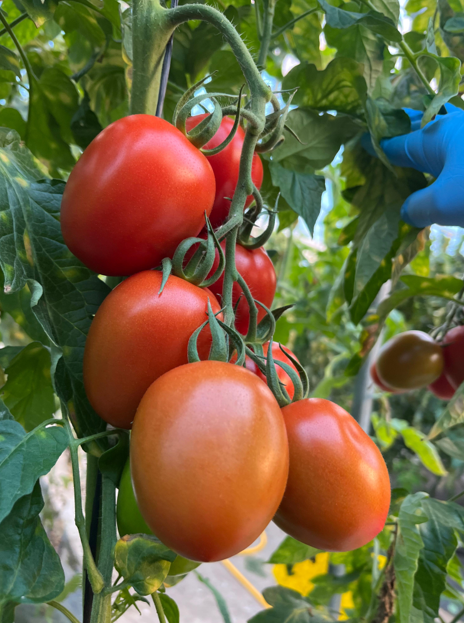} \centering
    \caption{Tomatoes}
    \label{fig:sub2}
  \end{subfigure}
  \caption{Crop and tomatoes fruit} \label{fig:Fig1}
\end{figure}

A central pathway, 2 meters wide, serves as the main access route and divides the greenhouse into two symmetrical sections. On the northern side, there are eleven aisles, each measuring 2 meters in width and 12.5 meters in length. On the southern side, the eleven aisles maintain the same width but extend to 22.5 meters. Secondary paths, each 1 meter wide, branch off from the central aisle to facilitate the navigation of mobile robotic units within the facility. Tomato plants (Lycopersicon esculentum) are cultivated in coir substrate bags arranged in rows oriented from north to south, following a terrain slope of 1\% (see Figure \ref{fig:sub1}). The tomato variety MargimenoRZ (Figure \ref{fig:sub1}), a pear-shaped type, was planted on August 26, 2024, as part of a long crop cycle extending until the end of May 2025. The crop was established using a planting density of 1.6 plants per square meter, allowing for optimal development and yield throughout the extended cultivation period.

\subsubsection{Camera RBG-D Intel Realsense D435i}

For image acquisition, the Intel RealSense D435i camera was used, an RGB-D sensor that integrates an active stereo vision system with a high-resolution RGB camera and an inertial mapping unit (IMU). This camera allows simultaneous color imaging (RGB format) and depth mapping, with a field of view of approximately 86° × 57° and a capture rate of up to 90 Hz, depending on the configured resolution. During the experiments, only the RGB images were used.

\begin{figure}[H]
    \centering
    \includegraphics[width=0.6\linewidth]{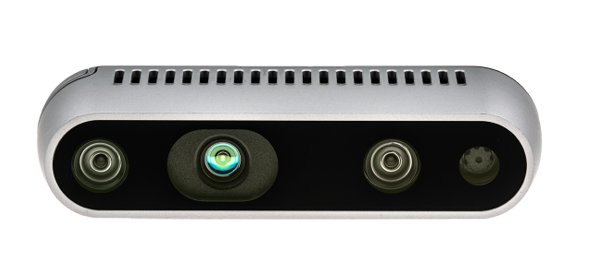}
    \caption{Intel RealSense D435i camera}
    \label{fig:placeholder}
\end{figure}

\subsubsection{Robot Operating Systems (ROS 2)}

For the development and integration of the robotic system, the Robot Operating System 2 (ROS2) \citep{macenski2022robot}, an open-source middleware widely adopted in research in service robotics, has been utilized. ROS 2 provides a modular and scalable infrastructure for design, simulation, control, and communication between heterogeneous robotic components. The modularity of ROS 2 has enabled the implementation of an extensible system with specialized nodes for hierarchical localization, geometric reconstruction, trajectory planning, and actuation. This provides a solid foundation for future developments and deployments in farming scenarios, driven by artificial intelligence and computer vision. In this work, ROS 2 facilitates camera interaction, data synchronization, and real-time processing.

\subsection{Methods}

\subsubsection{Hierarchical Localization (HLoc)}

For this work, Hierarchical Localization (HLoc) is used, an advanced approach that accurately estimates the position and orientation of a robot or autonomous agent in large-scale and complex environments. This technique combines the best of two worlds: the efficiency of global image retrieval methods and the accuracy of local matching algorithms \citep{sarlin2019from}. The process is divided into two main phases. 

In the first step, image retrieval is performed using image retrieval techniques, where a query image (e.g., a current capture from the robot's camera) is compared against a georeferenced database to identify a reduced set of candidate images that might correspond to similar locations. This step enables a drastic reduction in the search space, which is essential in large-scale scenarios. In the second phase, a local point correspondence (usually using descriptors such as SIFT, SuperPoint, or R2D2) is applied between the input image and the selected candidates. From these matches, the 6D pose of the agent is accurately estimated using techniques such as Perspective-n-Point (PnP) in combination with RANSAC to mitigate the effects of mismatches.

\subsubsection{Global Structure-from-Motion Revisited (GLOMAP)}

In combination with the above strategy, the Global Structure-from-Motion three-dimensional reconstruction technique (GLOMAP), a robust and scalable approach for estimating both scene geometry and camera trajectories from a set of unordered or partially ordered images, has also been utilized \citep{pan2024global}.

Unlike incremental methods, which reconstruct the scene by adding one image at a time and iteratively refining the solution, global approaches simultaneously estimate all camera poses from the relative relationships between pairs of images, commonly extracted by visual feature matching. In particular, GLOMAP relies on an initial exhaustive or hierarchical matching phase to find robust correspondences between images, followed by estimation of the relative rotation matrices using rotation averaging techniques. Subsequently, camera positions are resolved through translation averaging, thus completing the initial estimation of the complete camera trajectory. Once the trajectory is obtained, a dense or sparse reconstruction of the scene is performed using multi-view triangulation, complemented with global bundle adjustment to refine the geometry and reduce the reprojection error.

\section{Result and discussion}

In this section, the results obtained are analysed and discussed.

\subsection{Data adquisition}

For visual data acquisition, an Intel RealSense D435i camera was used, connected to a computer running Linux Ubuntu 22.04 and ROS 2 (Humble Hawksbill). ROS was used to capture a RGB video of the tomato crop from different positions accessible by a mobile robot, with a focus on the bunches. The recording was stored in rosbag format, which facilitates the synchronization and subsequent playback of the data acquired in ROS.

Once the capture was completed, a node was developed in ROS 2 for converting the sequence stored in rosbag (\texttt{.db3} extension with \texttt{metadata.yaml}) to individual images in \texttt{.png} format, extracted a total of 1600 frames at a frequency of 10 Hz. This processing enabled the generation of a folder containing the video frames, accompanied by a configuration file \texttt{pairs.txt}, which is necessary for executing the 3D reconstruction algorithm. In this file, the pairs of images to be compared are defined, considering exhaustive combinations between all the frames. This dataset, structured in images and comparison relations, constitutes the base input for the following phases of the three-dimensional reconstruction of the environment using Structure-from-Motion techniques.

\subsection{Model reconstruction with HLoc and GLOMAP.}

\begin{figure}[H]
    \centering
    \includegraphics[width=1\linewidth]{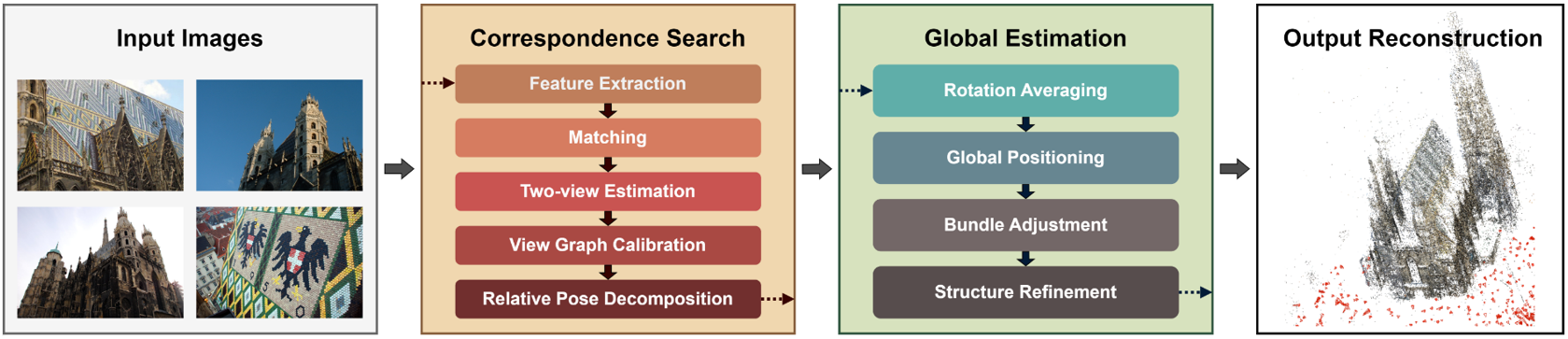}
    \caption{Pipeline of proposed GLOMAP system \citep{schonberger2016structure}}
    \label{4}
\end{figure}

The GLOMAP algorithm has as input a COLMAP dataset (Schonberger et al., 2016), which requires a single-view image dataset and a configuration file that relates the images. This technique follows the pipeline presented in Figure \ref{4}.

\begin{figure}[H]
    \centering
    \includegraphics[width=0.75\linewidth]{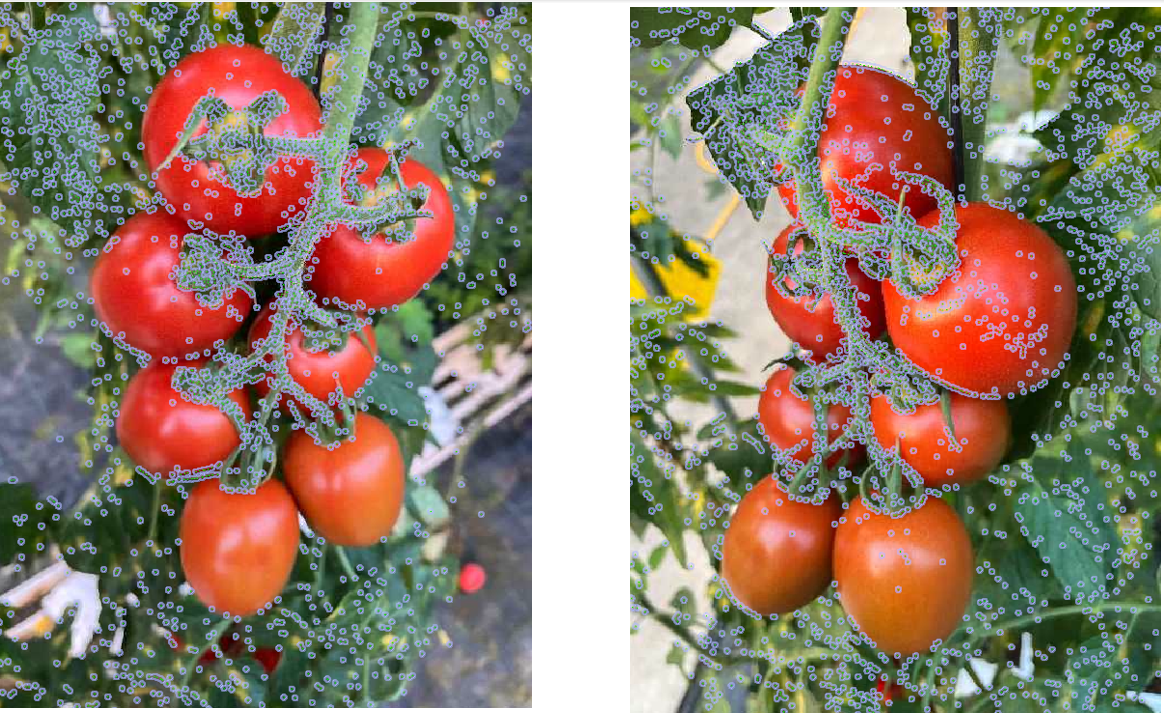}
    \caption{Matches found in the frames}
    \label{fig:placeholder6}
\end{figure}

Initially, the algorithm performs a matching of the images, locating the most important surf spots as ordered in file \texttt{pairs.txt}, as shown in Figure \ref{fig:placeholder6}.

\begin{figure}[H]
    \centering
    \includegraphics[width=1\linewidth]{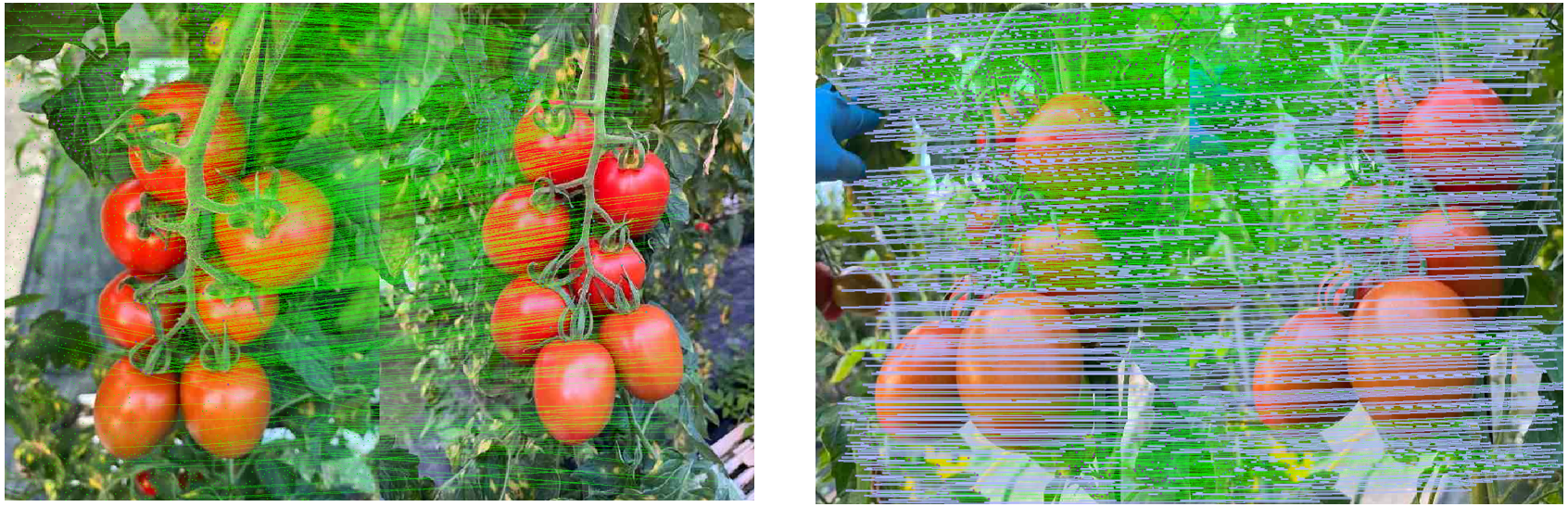}
    \caption{Matching found that coincide between different frames}
    \label{7}
\end{figure}

The process of estimating and decomposing the points begins. This is where HLoc comes in as it georeferences the analysed images, selecting the points that coincide and estimating the translation that has been carried out, as shown in Figure \ref{7}. This process can be observed in this video\footnote{\url{https://youtu.be/Uy_BpvbG7sc}}.

\begin{figure}[H]
    \centering
    \includegraphics[width=0.8\linewidth]{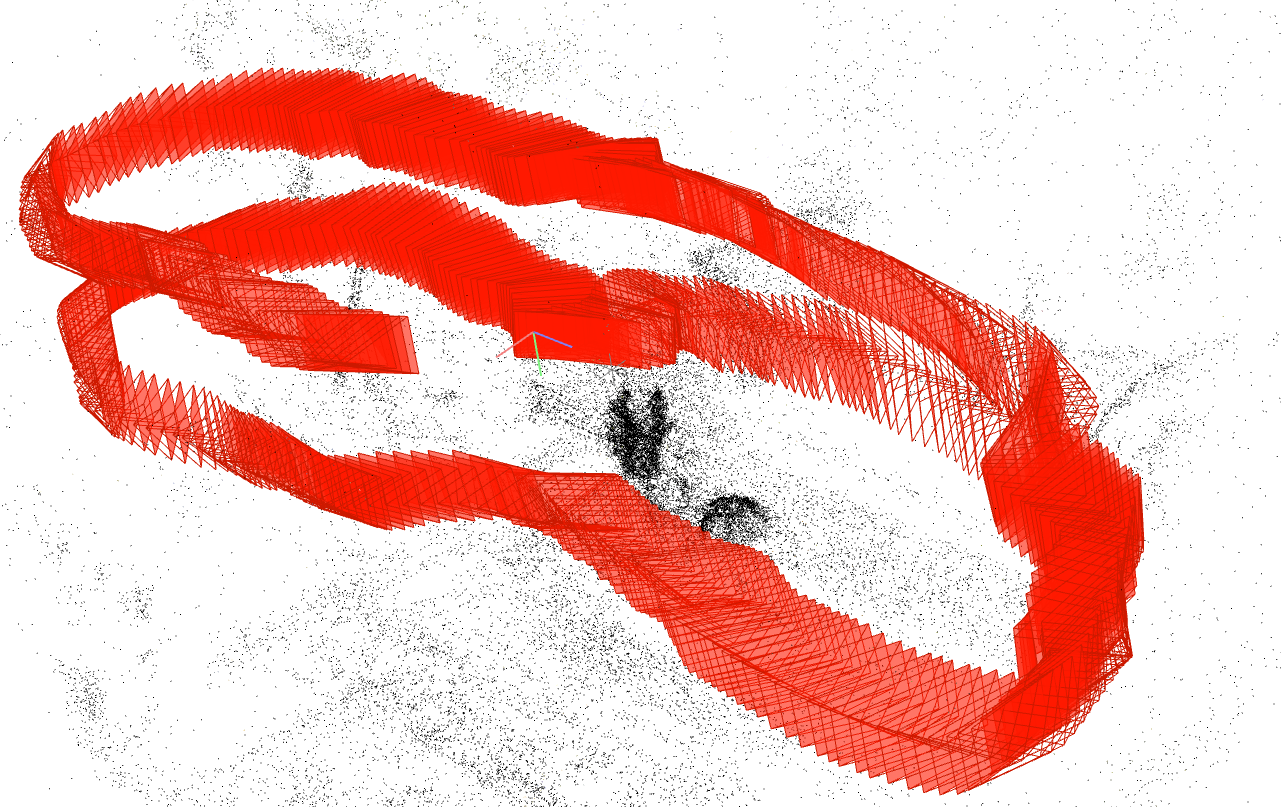}
    \caption{The trajectory shown in red and the GLOMAP point cloud in black.}
    \label{fig:placeholder7}
\end{figure}

The output of this algorithm generates a database.bin file that stores all the information obtained. This file, together with the set of images and named COLMAP, is the input to GLOMAP, which will perform the 3D reconstruction of the tomato cluster (see video\footnote{\url{https://youtu.be/jOgvrbCWjhs }}). In this case, the algorithm is based on the principle of Interactive Control Point (ICP) to estimate the position of each point, thus allowing to find a position in space and forming an object. Figure \ref{fig:placeholder7} shows the camera trajectory, as well as the resulting point cloud.

\begin{figure}[H]
    \centering
    \includegraphics[width=1\linewidth]{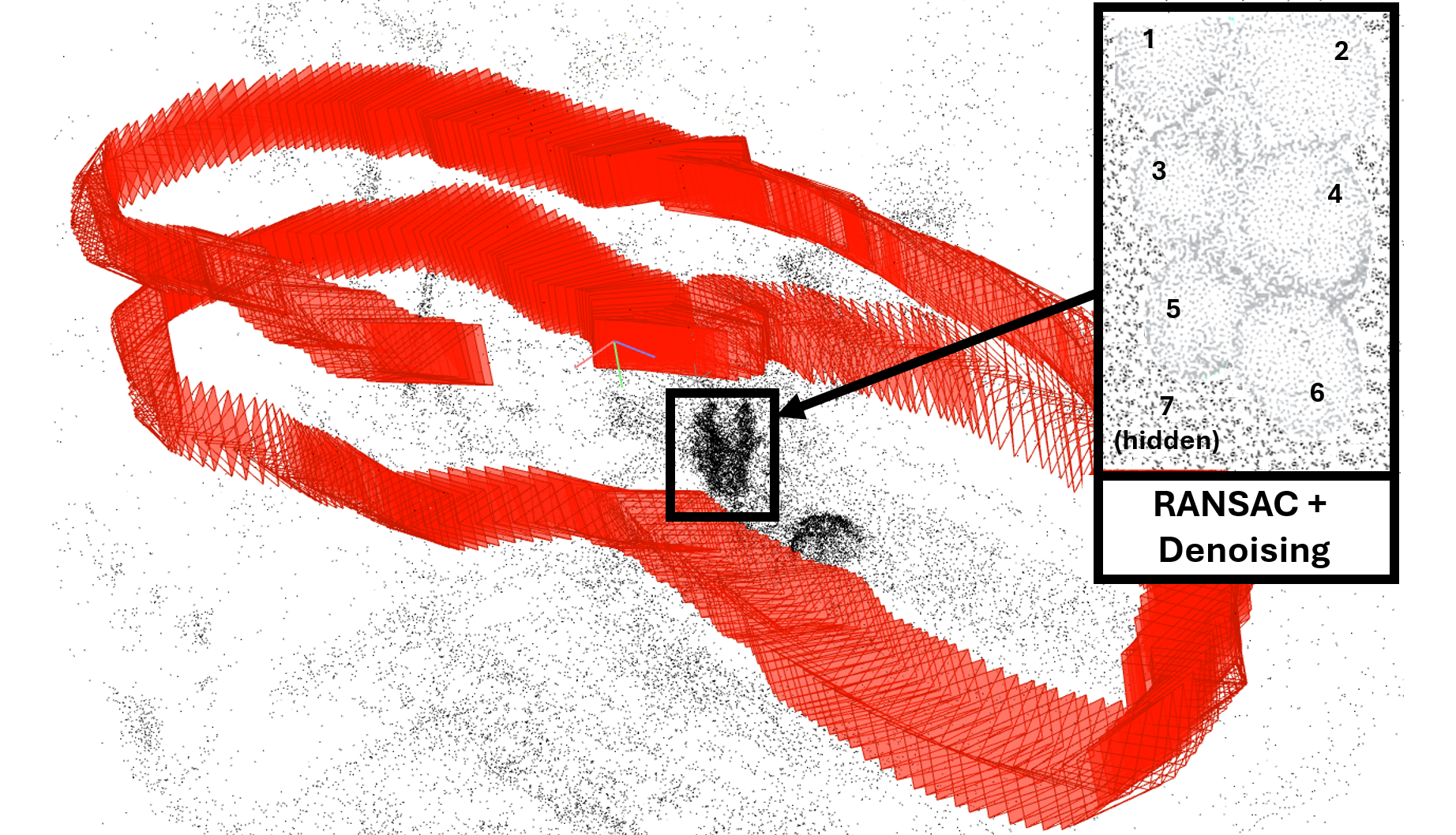}
    \caption{Reconstructed and treated model}
    \label{fig:placeholder9}
\end{figure}

As can be seen, in addition to the tomato cluster, spatial coordinates were recorded for other elements present in the greenhouse environment during the experimental day, including leaves, auxiliary structures, and crop components. To improve the quality of the acquired data, a two-stage cleaning process was applied. First, the Random Sample Consensus (RANSAC) algorithm \citep{barath2018graph} was employed to filter out points that exhibited greater dispersion or noise relative to the main surfaces. Subsequently, the CloudCompare tool (Girardeau-Montaut, 2016) was used to manually and accurately remove points close to the sensor that corresponded to visual artifacts or residual noise in the point cloud. This refinement allowed preserving only the elements of interest, resulting in the model shown in Figure~\ref{fig:placeholder9}. 

% \begin{figure}[H]
%     \centering
%     \includegraphics[width=1\linewidth]{image8.png}
%     \caption{Reconstructed and treated model}
%     \label{fig:placeholder8}
% \end{figure}

To validate the fidelity of this reconstruction, the resulting 3D model was compared against ground-truth measurements obtained manually on the real tomato cluster. Specifically, the diameter, centroid position, and orientation of each tomato were physically measured using a digital caliper and compared against the corresponding values extracted from the reconstructed point cloud. The real tomato exhibited an average diameter of approximately 5~cm ($\pm0.02$), with its centroid located at a distance of 2.6~cm ($\pm0.01$) from the main axis of the cluster. To determine the orientation and position, validation was carried out with regard to the junction of the two upper tomatoes on the main branch, yielding the values shown in Table \ref{tab:pose}.

\begin{table}[h]
\centering
\caption{Estimated 3D position and orientation (roll, pitch, yaw) of each tomato within the representative cluster.}
\label{tab:pose}
\begin{tabular}{@{}lcccccc@{}}
\toprule
\textbf{Tomato} & \textbf{x (mm)} & \textbf{y (mm)} & \textbf{z (mm)} & \textbf{Roll (°)} & \textbf{Pitch (°)} & \textbf{Yaw (°)} \\
\midrule
T1 & 12.3  & 8.1 & 210.5 & 2.1  & 3.0  & 5.4  \\
T2 & -18.7 & 6.4 & 195.2 & 3.8  & 5.5  & 8.2  \\
T3 & 22.4  & 5.9 & 193.8 & 4.5  & 4.2  & 6.9  \\
T4 & -25.1 & 4.2 & 170.6 & 6.2  & 7.8  & 11.3 \\
T5 & 27.9  & 3.8 & 168.4 & 5.7  & 6.1  & 9.7  \\
T6 & -14.6 & 2.1 & 145.9 & 8.9  & 9.3  & 14.1 \\
T7 & 8.2   & 1.5 & 118.3 & 10.4 & 11.5 & 16.8 \\
\bottomrule
\end{tabular}
\end{table}

These ground-truth measurements were compared against the corresponding values extracted from the reconstructed point cloud, yielding a diameter error of approximately $\pm0.2$ mm, a centroid position error of approximately $\pm0.4$ mm, and an orientation error of approximately $\pm0.75$$\degree$ in all the tomatoes. These results confirm that the reconstructed model closely reproduces the geometry of the real tomato cluster, both in fruit dimensions and spatial arrangement, supporting the reliability of the subsequent centroid and orientation estimation used to define the picking coordinate for the robotic gripper.

\section{Conclusion} \label{sec: 5}

The following conclusions can be drawn from the study. First, the use of a monocular RGB camera (Intel RealSense D435i) combined with ROS 2 enabled the acquisition and preprocessing of high-resolution image sequences in a structured and replicable manner, facilitating their use in advanced 3D reconstruction pipelines. Second, the integration of HLoc and GLOMAP within the Structure-from-Motion framework successfully reconstructed the three-dimensional morphology of tomato clusters, even under challenging visual conditions with occlusions and variable lighting. Third, the preprocessing steps, including the generation of \texttt{pairs.txt} for image pairing and the conversion of rosbag files to individual .png images, proved to be essential for optimizing the performance of the reconstruction algorithms. Fourth, a two-stage data refinement process, combining automatic filtering using RANSAC and manual denoising with CloudCompare, significantly improved the quality of the resulting point cloud, preserving the relevant structures while eliminating visual artifacts. Furthermore, validation against manual ground-truth measurements confirmed the geometric fidelity of the reconstructed models, with errors of only a few millimeters in fruit diameter and centroid position and a few degrees in orientation, supporting the reliability of the estimated picking coordinates for robotic manipulation. Finally, the proposed methodology demonstrates that accurate 3D modeling of agricultural elements can be achieved with low-cost monocular vision systems, which reduces hardware requirements and enhances the scalability of robotic perception solutions in commercial greenhouse environments.

\section*{Acknowledgments} 
This work has been carried out within the framework of the LIFE-ACCLI\\AMTE project (LIFE23-CCAES-LIFE-ACCLIMATE/101157315), and a Er-\\asmus+ Mobility grants from CeiA3. The first author, Fernando Cañadas-Aránega, holds an FPI grant (PRE2022-102415) from the Spanish Ministry of Science, Innovation, and Universities..
%% Loading bibliography style file
%\bibliographystyle{model1-num-names}
\bibliographystyle{cas-model2-names}

% Loading bibliography database
\bibliography{cas-refs}

@inproceedings{aguilar2024preliminary,
  author    = {Aguilar, F. J. and Blanco, J. L. and Nemmaoui, A. and Ca{\~n}adas-Ar{\'a}nega, F. and Aguilar, M. and Moreno, J. C.},
  title     = {Preliminary results of a low-cost portable terrestrial lidar based on {ICP-SLAM} algorithms application to automatic forest digital inventory},
  booktitle = {6th Euro-Mediterranean Conference for Environmental Integration (EMCEI-2024)},
  address   = {Marrakech, Morocco},
  pages     = {15--18},
  year      = {2024}
}

@article{bachche2015deliberation,
  author  = {Bachche, S.},
  title   = {Deliberation on design strategies of automatic harvesting systems: A survey},
  journal = {Robotics},
  volume  = {4},
  number  = {2},
  pages   = {194--222},
  year    = {2015}
}

@inproceedings{barath2018graph,
  author    = {Barath, D. and Matas, J.},
  title     = {Graph-cut {RANSAC}},
  booktitle = {Proceedings of the IEEE Conference on Computer Vision and Pattern Recognition (CVPR)},
  pages     = {6733--6741},
  year      = {2018}
}

@article{moreno2022modelado,
  author = {Moreno {\'U}beda, J. C. and Ca{\~n}adas-Ar{\'a}nega, F. and Rodr{\'i}guez, F. and S{\'a}nchez-Hermosilla, J. and Gim{\'e}nez, A.},
  title  = {3D modelling and design of a collaborative robot for transport tasks in greenhouses},
  journal={XVLI Jornadas de Autom{\'a}tica, Málaga, Andalusia, Spain.},
  volume  = {43},
  year   = {2022}
}

@article{canadas2024multimodal,
  author  = {Ca{\~n}adas-Ar{\'a}nega, F. and Blanco-Claraco, J. L. and Moreno, J. C. and Rodriguez-Diaz, F.},
  title   = {Multimodal mobile robotic dataset for a typical Mediterranean greenhouse: The {GreenBot} dataset},
  journal = {Sensors},
  volume  = {24(6)},
  number  = {6},
  pages   = {1897},
  year    = {2024}
}

@article{canadas2024autonomous,
  author  = {Ca{\~n}adas-Ar{\'a}nega, F. and Moreno, J. C. and Blanco-Claraco, J. L. and Gim{\'e}nez, A. and Rodr{\'i}guez, F. and S{\'a}nchez-Hermosilla, J.},
  title   = {Autonomous collaborative mobile robot for greenhouses: Design, development, and validation tests},
  journal = {Smart Agricultural Technology},
  volume  = {9},
  pages   = {100606},
  year    = {2024}
}

@article{canadas2026adaptive,
  author  = {Ca{\~n}adas-Ar{\'a}nega, F. and Wollherr, D. and Guzm{\'a}n, J. L. and Moreno, J. C. and Blanco-Claraco, J. L.},
  title   = {An adaptive control architecture for slope and terrain compensation in autonomous navigation in Mediterranean greenhouses},
  journal = {arXiv preprint arXiv:2609.02487},
  year    = {2026}
}

@article{canadas2026integrated,
  author  = {Ca{\~n}adas-Ar{\'a}nega, F. and Mu{\~n}oz, M. and Moreno, J. C. and Blanco-Claraco, J. L.},
  title   = {Integrated cloud-based architecture for robot-robot and human-robot collaboration using {ROS 2--MQTT} in Mediterranean Greenhouses},
  journal = {arXiv preprint arXiv:2606.22682},
  year    = {2026}
}

@article{canadas2026greenseg,
  author  = {Ca{\~n}adas-Ar{\'a}nega, F. and Moreno, J. C. and Blanco-Claraco, J. L.},
  title   = {{GreenSeg}: Ground Segmentation Algorithm for Agricultural Robots in Mediterranean Greenhouses using {RGB-D} Point Clouds},
  journal = {arXiv preprint arXiv:2605.25279},
  year    = {2026}
}

@article{canadas2026ros2,
  author  = {Ca{\~n}adas-Ar{\'a}nega, F. and Ma{\~n}as-{\'A}lvarez, F. J. and Moreno, J. C. and Blanco-Claraco, J. L.},
  title   = {A {ROS2} Benchmarking Framework for Hierarchical Control Strategies in Mobile Robots for Mediterranean Greenhouses},
  journal = {arXiv preprint arXiv:2602.15162},
  year    = {2026}
}

@inproceedings{feng2015design,
  author    = {Feng, Q. and Wang, X. and Wang, G. and Li, Z.},
  title     = {Design and test of tomatoes harvesting robot},
  booktitle = {2015 IEEE International Conference on Information and Automation},
  pages     = {949--952},
  year      = {2015}
}

@article{ge2021yolox,
  author  = {Ge, Z. and Liu, S. and Wang, F. and Li, Z. and Sun, J.},
  title   = {{YOLOX}: Exceeding {YOLO} series in 2021},
  journal = {arXiv preprint arXiv:2107.08430},
  year    = {2021}
}

@inproceedings{pan2024global,
  author    = {Pan, L. and Bar{\'a}th, D. and Pollefeys, M. and Sch{\"o}nberger, J. L.},
  title     = {Global Structure-from-Motion Revisited},
  booktitle = {European Conference on Computer Vision (ECCV)},
  pages     = {58--77},
  publisher = {Springer Nature Switzerland},
  address   = {Cham},
  year      = {2024}
}

@article{rong2022fruit,
  author  = {Rong, J. and Wang, P. and Wang, T. and Hu, L. and Yuan, T.},
  title   = {Fruit pose recognition and directional orderly grasping strategies for tomato harvesting robots},
  journal = {Computers and Electronics in Agriculture},
  volume  = {202},
  pages   = {107430},
  year    = {2022}
}

@inproceedings{sarlin2019from,
  author    = {Sarlin, P. E. and Cadena, C. and Siegwart, R. and Dymczyk, M.},
  title     = {From coarse to fine: Robust hierarchical localization at large scale},
  booktitle = {Proceedings of the IEEE/CVF Conference on Computer Vision and Pattern Recognition (CVPR)},
  pages     = {12716--12725},
  year      = {2019}
}

@inproceedings{schonberger2016structure,
  author    = {Sch{\"o}nberger, J. L. and Frahm, J. M.},
  title     = {Structure-from-motion revisited},
  booktitle = {Proceedings of the IEEE Conference on Computer Vision and Pattern Recognition (CVPR)},
  pages     = {4104--4113},
  year      = {2016}
}

@article{aranega2025nbv,
  title={NBV Planning for the detection of hidden tomatoes in greenhouses with AgriSEE},
  author={Ca{\~n}adas-Aranega, Fernando and Border, Rowan and Blanco, Jose Luis and Moreno, Jose Carlos},
  journal={XVLI Jornadas de Autom{\'a}tica, Málaga, Andalusia, Spai. },
  volume={46(2)},
  year={2025}
}

@article{macenski2022robot,
  title={Robot operating system 2: Design, architecture, and uses in the wild},
  author={Macenski, Steven and Foote, Tully and Gerkey, Brian and Lalancette, Chris and Woodall, William},
  journal={Science robotics},
  volume={7},
  number={66},
  pages={eabm6074},
  year={2022},
  publisher={American Association for the Advancement of Science}
}

@inproceedings{taqi2017cherry,
  author    = {Taqi, F. and Al-Langawi, F. and Abdulraheem, H. and El-Abd, M.},
  title     = {A cherry-tomato harvesting robot},
  booktitle = {2017 18th International Conference on Advanced Robotics (ICAR)},
  pages     = {463--468},
  year      = {2017}
}

@misc{vela2025hloc,
  author       = {Vela, P.},
  title        = {hloc-{GLOMAP}: Visual {SLAM} pipeline with Hierarchical Localization and {GLOMAP}},
  howpublished = {\url{https://github.com/pablovela5620/hloc-glomap}},
  year         = {2025}
}

@article{zheng2024fruit,
  author  = {Zheng, X. and Rong, J. and Zhang, Z. and Yang, Y. and Li, W. and Yuan, T.},
  title   = {Fruit growing direction recognition and nesting grasping strategies for tomato harvesting robots},
  journal = {Journal of Field Robotics},
  volume  = {41},
  number  = {2},
  pages   = {300--313},
  year    = {2024}
}

% Biography

%\bio{}
% Here goes the biography details.
%\endbio

%\bio{pic1}
% Here goes the biography details.
%\endbio

\end{document}